\documentclass{article}

\usepackage{PRIMEarxiv}

\usepackage[utf8]{inputenc} 
\usepackage[T1]{fontenc}    
\usepackage{hyperref}       
\usepackage{url}            
\usepackage{booktabs, dcolumn}       
\usepackage{amsfonts}       
\usepackage{nicefrac}       
\usepackage{microtype}      
\usepackage{lipsum}
\usepackage{fancyhdr}       
\usepackage{graphicx}       

\usepackage{subcaption}
\usepackage{multirow}
\usepackage{mathtools}
\usepackage{amsmath}

\usepackage{subcaption}
\usepackage{amsmath}

\usepackage{nicefrac}
\usepackage{multirow}
\usepackage{booktabs, dcolumn}
\usepackage{mathtools}

\usepackage{microtype}

\graphicspath{{media/}}     

\title{Tackling the Class Imbalance Problem of Deep Learning Based Head and Neck Organ Segmentation}

\author{
  Elias Tappeiner, Martin Welk, Rainer Schubert \\
  Department for Biomedical Computer Science and Mechatronics \\
  UMIT -- Private University for Health Sciences, Medical Informatics and Technology \\
  Hall in Tyrol\\
  \texttt{elias.tappeiner@umit-tirol.at} \\
}

\begin{document}
\maketitle

\begin{abstract}
    \textbf{Purpose:} The segmentation of organs at risk (OAR) is a required precondition for the cancer treatment with image guided radiation therapy. The automation of the segmentation task is therefore of high clinical relevance. Deep Learning (DL) based medical image segmentation is currently the most successful approach, but suffers from the over-presence of the background class and the anatomically given organ size difference, which is most severe in the head and neck (HAN) area. 

    \textbf{Methods:} To tackle the HAN area specific class imbalance problem we first optimize the patch-size of the currently best performing general purpose segmentation framework, the nnU-Net, based on the introduced class imbalance measurement, and second, introduce the class adaptive Dice loss to further compensate for the highly imbalanced setting. 

    \textbf{Results:} Both the patch-size and the loss function are parameters with direct influence on the class imbalance and their optimization leads to a 3\% increase of the Dice score and 22\% reduction of the 95\% Hausdorff distance compared to the baseline, finally reaching $0.8\pm0.15$ and $3.17\pm1.7$ mm for the segmentation of seven HAN organs using a single and simple neural network. 

    \textbf{Conclusion:} The patch-size optimization and the class adaptive Dice loss are both simply integrable in current DL based segmentation approaches and allow to increase the performance for class imbalanced segmentation tasks.
\end{abstract}

\keywords{Deep Learning, Segemntation, Head and Neck, Class Imablance, Radiation Therapy}

\section{Introduction}
Cancer is after cardiovascular diseases the second most common cause of death. Among the newly diagnosed cancer incidences, statistically 3\% are tumors of the head and neck (HAN) region~\cite{Siegel21}. Due to the complex anatomy of the area, characterized by a large number of small soft tissue organs, image guided radiotherapy is the primary choice of treatment for HAN cancer. The segmentation of the organs at risk (OAR) on the planning CT scans is necessary for the radiotherapy and the main reason of treatment delivery delays throughout the clinical pathway of the therapy. The segmentation is time consuming, requires several highly educated medical experts and is still mainly performed manually, further observer variations are well-documented~\cite{Nikolov2018}. Due to the time-consuming and subjective manual process a field of research has developed around the automated segmentation of the HAN organs on medical images, with deep learning (DL) being the dominant and most successful learning based approach~\cite{Vrtovec2020}. Segmentation with DL can be interpreted as a voxel-wise classification problem using fully convolutional neural networks. The large difference in size of classes to be segmented can be defined as the class imbalance problem. Since the first introduction of a DL based multi-organ HAN segmentation approach~\cite{Tappeiner2019}, it is known that the HAN area is specially affected by the class imbalance problem. In addition to the large difference in ratio of background and foreground voxels, the HAN area is characterized by large size differences between the foreground classes themselves, which is anatomically given through the differently sized organs to be segmented. As a result the class imbalance causes a large performance difference in the segmentation of large and small organs~\cite{Vrtovec2020}. 

In this work we focus on the training window or patch-size as the hyper-parameter with a direct influence on the class imbalance, as most segmentation networks are, due to GPU memory constraints, trained with randomly sampled patches of the original 3D image. Hence, we introduce a measurement for the class imbalance of differently sized training patches and optimize the patch-size accordingly. Additionally, we adapt the classical multi-class Dice loss formulation which does not account for missing classes within patches. Our class adaptive Dice loss formulation is robust against missing classes, which is relevant for sparse class distributions within the image dataset and for the training with smaller patch-sizes. We incorporate both, the class imbalance optimized patches and the class adaptive Dice loss into the currently best performing general purpose segmentation approach, the nnU-Net framework~\cite{Isensee2021}, and are able to increase the performance of its baseline version. The introduced multi-class confidence analysis following the work of Li et al.~\cite{Li2021} also reveals an increased segmentation confidence for mid-sized organs due to the class label imbalance optimized patch-size.

\section{Related Work}
Guo et al.~\cite{Guo2020} and Gao et al.~\cite{Gao2019} were the first to specifically address and solve the class imbalance problem of the HAN area by using several different cascaded networks. The approaches are inspired by the work of clinical experts, first segmenting large and easy anchor organs and then zooming in to segment the harder small soft tissue organs. Similarly, the authors combined a strong large organ segmentation network, a small organ localization network and specific small organ segmentation networks effectively reducing the class imbalance of each network. In their followup work, the FocusNetv2, Gao et al.~\cite{Gao2020} further incorporated autoencoder based shape priors~\cite{Oktay2018} and adversarial training~\cite{Goodfellow2014} into the small organ networks, achieving a Dice score (DSC) of 0.84 and a 95\% Hausdorff distance (95HD) of 2.17 mm which are the currently best reported results on the MICCAI 2015 HAN segmentation challenge reference dataset~\cite{Raudaschl2017}. An implicit reduction of the class imbalance, especially in favor of the small organs that are often visible in just a few CT slices, is recently achieved by hybrid networks using 2D convolutions and 3D convolutions in their architecture. Chen et al.~\cite{Chen2021} used 2D convolutions for the extraction of fine edges and 3D convolutions for coarse and fine semantic features in a single UNet~\cite{Ronneberger2015} based architecture. Tang et al.~\cite{Tang2021} extended a 2D UNet with an additional 3D convolution based context-aware attention path and were able to achieve state-of-the-art using a single HAN organ segmentation network. 

Differently to architectural changes of the network, adapted cost functions can also reduce the class imbalance problem of DL. Roth et al.~\cite{Roth2017} presented the first DL based multi-organ segmentation approach of the abdominal area and applied a class weighted cross entropy (CE) loss function. The CE is an information theoretical measurement for probability distribution differences and allows to calculate the difference between the network's voxel-wise class prediction and the ground truth. As the CE is the classical loss function for image classification, Milletari et al.~\cite{Milletari2016} proposed the DSC as a volume based overlap measurement to be used as a loss function for image segmentation. The Dice loss transforms the voxel-wise measurement into a semantic label overlap measurement and has become the state-of-the-art loss function of the field. Effectively reducing the number of measurements to the number of labels, the Dice loss also reduces the sensitivity of the loss regarding the class imbalance effect. However, the Dice loss is not able to eliminate the problem due to its intrinsic bias towards large volumes~\cite{Nikolov2018} as well as the remaining severe over-presence of the largest class during training. Consequently, Carole et al.~\cite{Sudre2017} introduced the generalized Dice score (GDSC), which adaptively weights the DSC by the current class size. However, in a previous work~\cite{Tappeiner2019} we showed that the GDSC introduces noise in the learning curve by the adaptive weights and missing classes in case of the common patch-based training setting. Zhu et al.~\cite{Zhu2019} investigated different loss functions specifically for the imbalanced HAN area and showed the combination of the Dice loss and the focal loss~\cite{Lin2020} to outperform the plain Dice loss. Isensee et al.~\cite{Isensee2021} proposed to combine the CE and dice loss to measure both the voxel-wise class predictions and the semantic label overlap and were able to show advancements in many different segmentation tasks using the combined loss function in their nnU-Net. 

Another approach to analyze the class imbalance in neural networks for image segmentation is presented by Li et al.~\cite{Li2021}. The authors found that the network output of under-represented classes tend to shift towards the decision boundary during test time, whereas well-represented classes are unaffected. As a result the authors claim that an overfitting of the small sized classes occurs during the training. 
For their analysis of the class imbalance induced overfitting, the authors suggest to plot the logit output of the training data against the test data, which we adapt and confirm for our given multi-class setting.
\bigskip


\section{Method}
\label{li:sec:method}
\subsection{Dataset}
For our study of the class imbalance problem in the HAN area we utilize the MICCAI 2015 HAN auto segmentation challenge dataset~\cite{Raudaschl2017}. The CT images of the dataset are from the $0522$ multi-institutional clinical study of the Radiation Therapy Oncology Group~\cite{Ang2014}, which made the data publicly available. The study contained multiple images of 111 patients with HAN cancer of the oropharynx, the hypopharynx or the larynx. The challenge dataset includes $40$ patient CT scans, with manual reference segmentations of nine structures: the left and right Parotid Gland (PG), the left and right Submandibular Gland (SG), the Optic Chiasm (OC), the Brainstem (BS), the Mandible (MA) and the left and right Optic Nerves (ON). Although the original images of the $0522$ study contained OAR reference segmentations for the radiotherapy planning, no standardized segmentation protocols existed at the time of the study and the segmented structures showed considerable differences in contouring. Accordingly, the nine organs for the dataset creation are iteratively re-contoured according to current scientific standard protocols until all segmentation experts agree and the observer bias is eliminated. For the scope of the challenge 25 specific images identified by their file names are released as training images, 10 as an offsite test set and the last 5 as an additional test set for the onsite event of the challenge. In our work we follow the challenge protocol regarding the dataset splits and combine the off- and onsite test images to one test set for our result presentation. The Submandibular Glands are not considered in our work as not all 40 CT scans contain the corresponding reference data.

\subsection{Segmentation Network Design}
Our work is based on the 3D nnU-Net framework of Isensee et al.~\cite{Isensee2021}. The authors claimed and showed that a well-parameterized UNet~\cite{Ronneberger2015} is hard to beat for any segmentation task and accordingly defined a set of well-proven fixed parameters and additional dataset dependent rule based parameters for a dynamically deep UNet. The fixed parameters are the learning rate, the optimizer, the data augmentation, the number of training iterations, the patch sampling strategy, the loss function, the inference using a sliding window approach and the post-processing as a largest component analysis. The most relevant dataset dependent parameters are the spacing and the patch-size further defining the UNet architecture. The spacing is evaluated as the median of the dataset in-plane spacing and the 10\textsuperscript{th} percentile of the out-plane spacing resulting in a spacing of $0.98\times0.98\times2.5$ mm. The patch-size is initialized to the dataset median after resampling and iteratively enlarged, simultaneously with the depth of the UNet to fill the available GPU memory using a fixed batch size of two resulting in a patch-size of $192\times160\times56$. The skeleton UNet is a basic UNet with two blocks of convolution, instance normalization~\cite{Ulyanov2016} and nonlinearity in each resolution, starting with a channel size of 32, which is getting doubled (halved) with each downsampling (upsampling) operation. 
To inject gradients deeper in the network, deep supervision with auxiliary losses are used for the upsampling layer of the encoder. For further details regarding the original 3D version of the nnU-Net we refer to the work of Isensee et al.~\cite{Isensee2021}.

\begin{figure}
    \centering
    \captionsetup[subfigure]{labelformat=empty}
    \begin{subfigure}{.4\linewidth}
        \centering
        \includegraphics[width=0.99\textwidth]{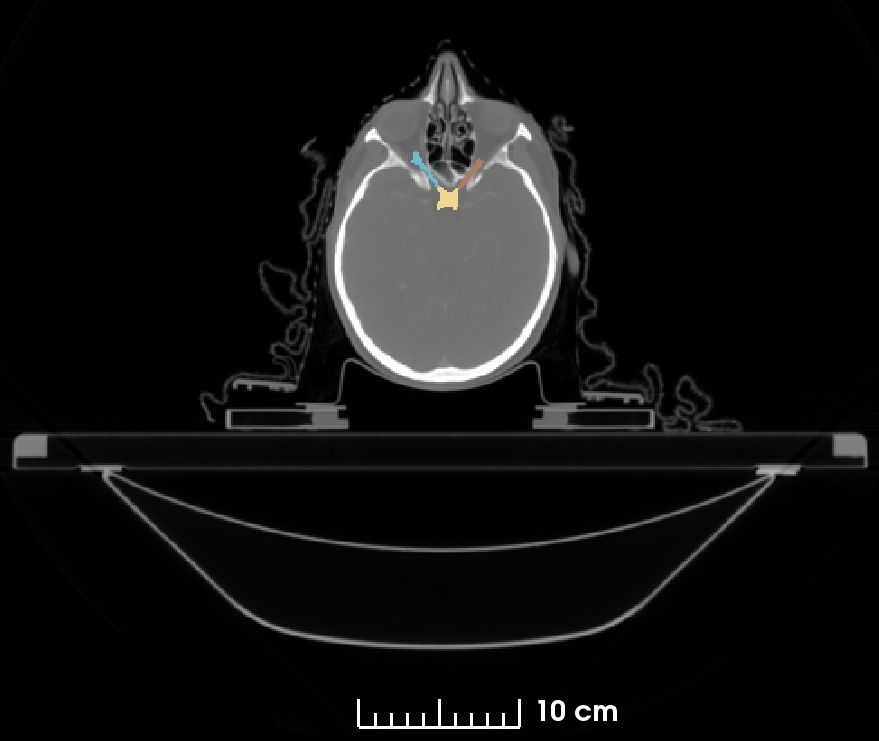}
    \end{subfigure} \hspace{.05\linewidth}
    \begin{subfigure}{.4\linewidth}
        \centering
        \includegraphics[width=0.99\textwidth]{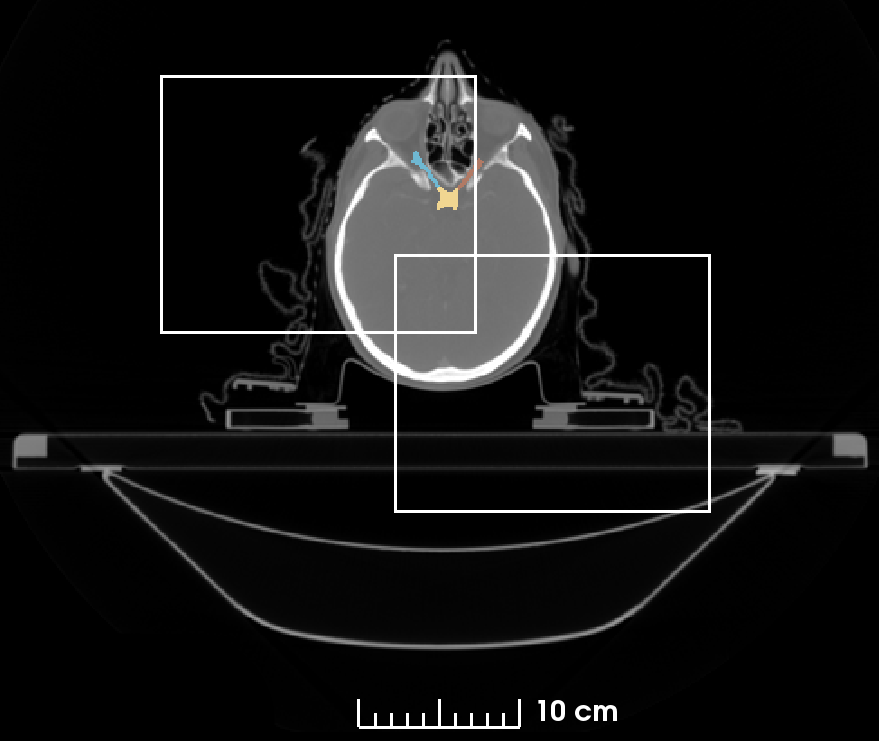}
    \end{subfigure} \\[1ex]
    \begin{subfigure}{.4\linewidth}
        \centering
        \includegraphics[width=0.99\textwidth]{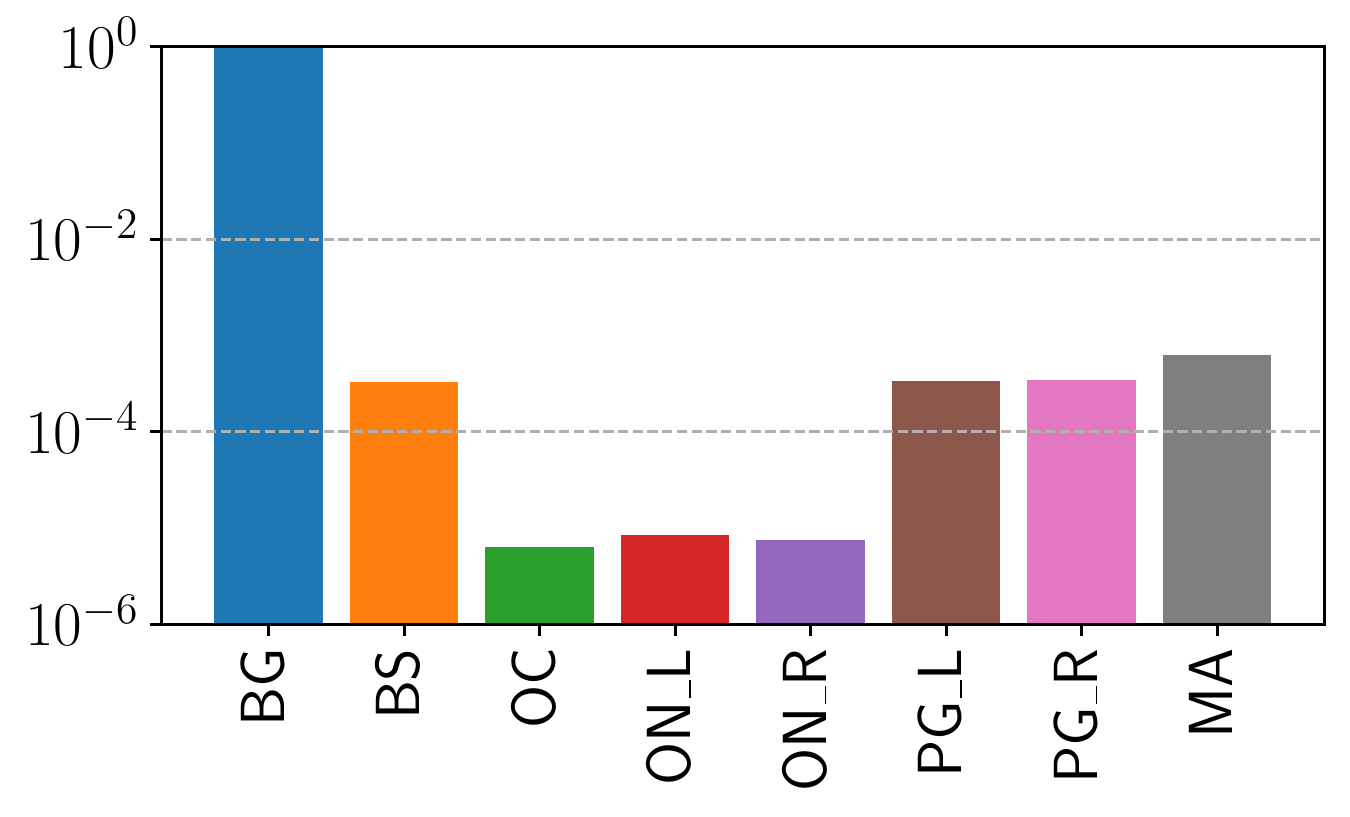}
        \caption{full volume (max patch-size)}
    \end{subfigure} \hspace{.05\linewidth}
    \begin{subfigure}{.4\linewidth}
        \centering
        \includegraphics[width=0.99\textwidth]{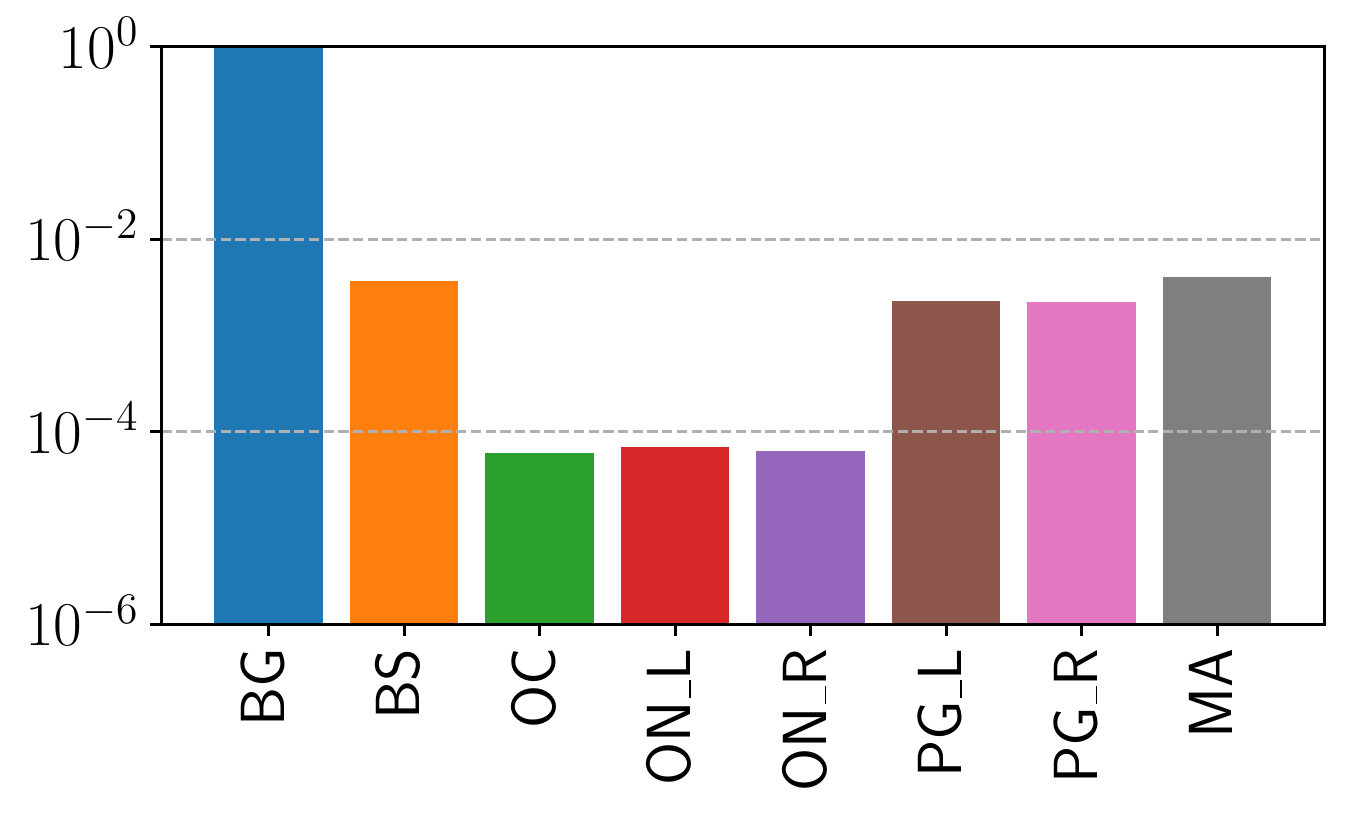}
        \caption{$192\times160\times56$ (large patch-size)}
    \end{subfigure} \\[3ex]
    \begin{subfigure}{.4\linewidth}
        \centering
        \includegraphics[width=0.99\textwidth]{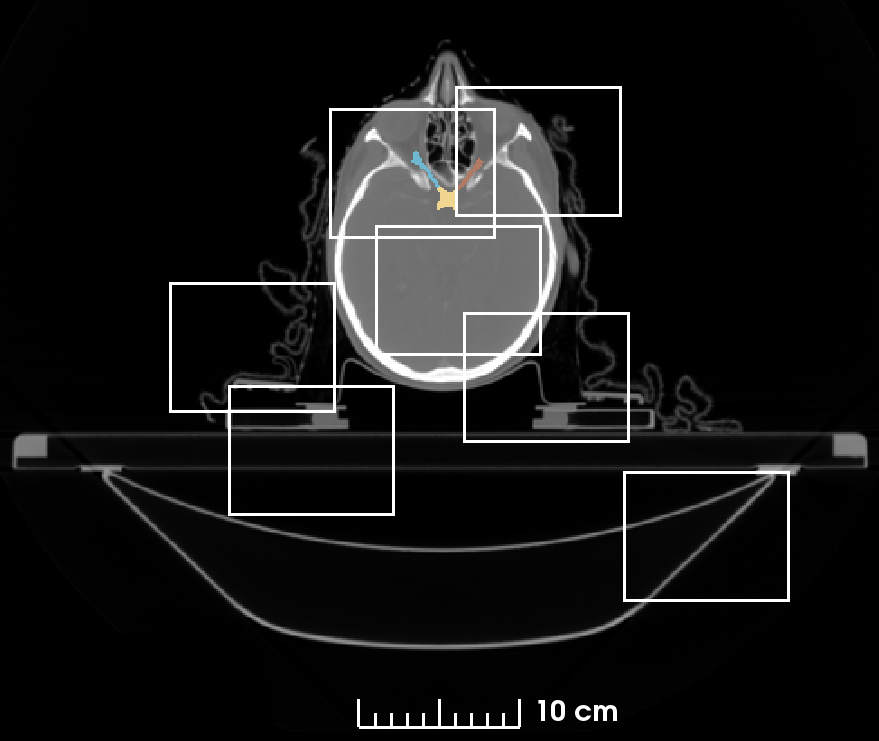}
    \end{subfigure} \hspace{.05\linewidth}
    \begin{subfigure}{.4\linewidth}
        \centering
        \includegraphics[width=0.99\textwidth]{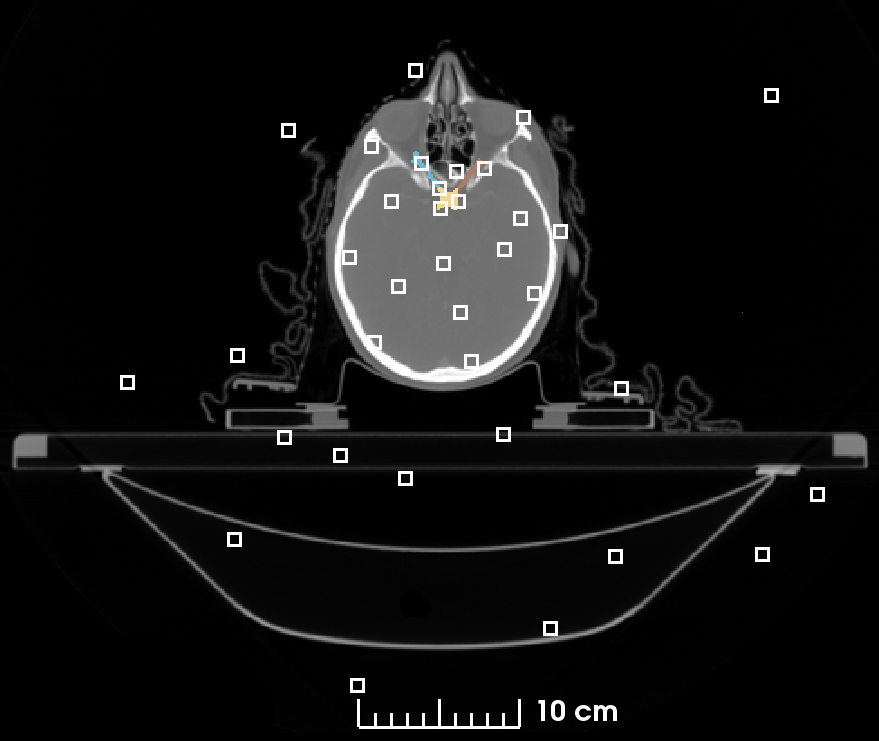}
    \end{subfigure} \\[1ex]
    \begin{subfigure}{.4\linewidth}
        \centering
        \includegraphics[width=0.99\textwidth]{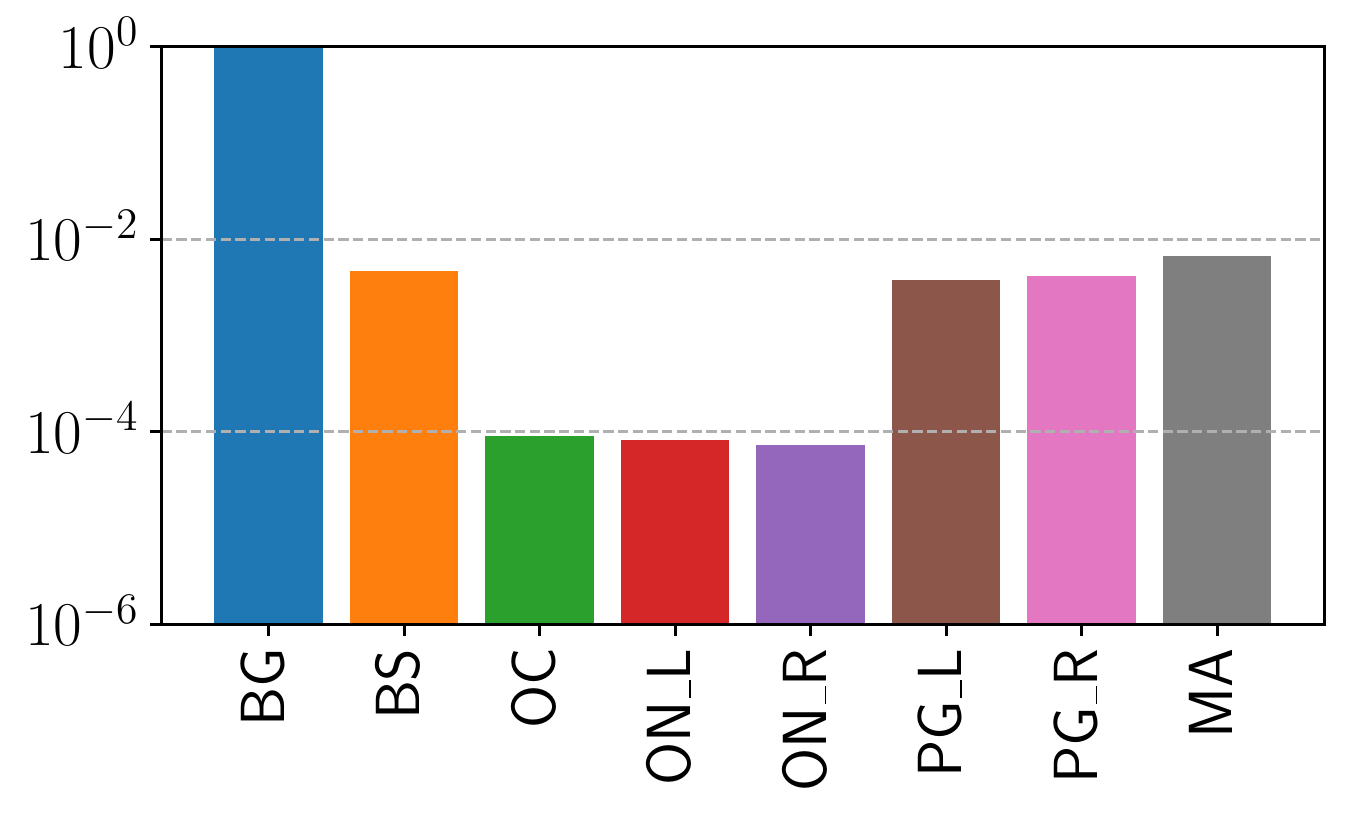}
        \caption{$96\times80\times48$ (small patch-size)}
    \end{subfigure} \hspace{.05\linewidth}
    \begin{subfigure}{.4\linewidth}
        \centering
        \includegraphics[width=0.99\textwidth]{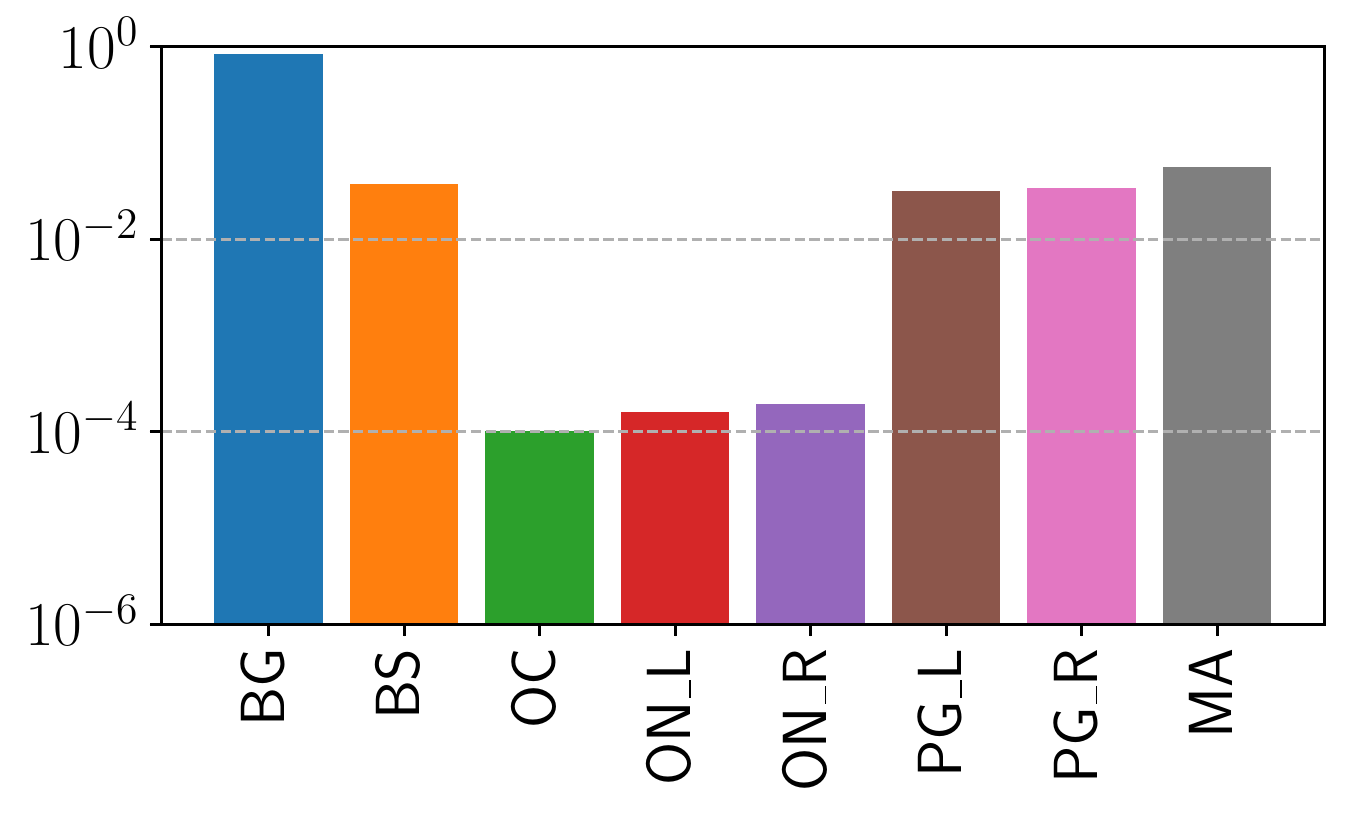}
        \caption{$8\times8\times8$ (min patch-size)}
    \end{subfigure}
    \caption{Average background and organ volume imbalance ratios of seven HAN organs (from left to right: Background, Brainstem, Optic Chiasm, Optic Nerve left/right, Parotid Gland left/right, Mandible) for four different patch-size sampling strategies, evaluated over one training epoch.}
    \label{li:fig:li}
\end{figure}

\subsection{Class Imbalance Measurement}
As the currently most advanced general purpose approach for medical image segmentation we mainly follow the 3D nnU-Net framework, but adapt the loss function and also the patch-size based on our class imbalance measurement as the parameters directly influencing the class imbalance while training. Figure~\ref{li:fig:li} shows the average imbalance of the organ and background volume ratios of the dataset within a training epoch for different training strategies. For the ratio measurement the dataset is rescaled following the spacing definition of the nnU-Net. Although the histograms visually show the difference of the organ volume ratios for the presented patch-size strategies, we propose to use the standard deviation $\sigma$ of the class ratios as a single measurement for the class imbalance. The standard deviation of the averaged in-patch organ ratios is a single and easily interpretable value. The ratios sum up to one, accordingly the standard deviation is the average distance to an ideally uniform distribution of in-patch organ ratios. Utilizing $\sigma$ as a cost function with the patch-size as parameter allows us to find the training parameter with a minimal imbalance for the given dataset.

\newpage

\subsection{Class Adaptive Dice Loss} 
\label{li:sec:la_dice_loss}
The loss function proposed by the nnU-Net is the CE+Dice loss combining probabilistic voxel-wise class predictions and label overlap measurements, which is also advised by the currently largest study of loss functions for medical image segmentation by Ma et al.~\cite{Ma2021}. The CE loss is used in its basic multi-class formulation as:
\begin{equation}
    \text{CE}(P,G) \; = \frac{1}{B} - \sum_{b,c,v} G_{bcv} \log(P_{bcv}) \;,
    \label{eq:celoss}
\end{equation}
with $P$ and $G$ being the one-hot-encoded prediction and ground truth volumes, consisting of $B$ batches, $C$ classes and $V$ voxels. The multi-class Dice loss using $\epsilon$ as a small value for numeric stability is defined as:
\begin{equation}
	\text{Dice}(P,G)=\frac{1}{BC}\sum_{b,c} \frac{2 \sum_{v}P_{bcv}G_{bcv}+\epsilon}{\sum_{v}P_{bcv}+\sum_{v}G_{bcv}+\epsilon} \; .
	\label{eq:diceloss}
\end{equation}
The Dice loss formulation of the nnU-Net follows the batch Dice loss of Kodym et al.~\cite{Kodym2018} with the adaptation of ignoring the background class. Contrary to the original Dice definition, Kodym et al. propose to evaluate the DSC with the batch as part of the volume instead of averaging the DSC over the batches. Accordingly, the Dice loss formulation used in the nnU-Net is given by:
\begin{equation}
    \text{nnU-Dice}(P,G)=\frac{1}{C-1}\sum_{c-1} \frac{2 \sum_{b,v}P_{bcv}G_{bcv}+\epsilon}{\sum_{b,v}P_{bcv}+\sum_{b,v}G_{bcv}+\epsilon} \;.
    \label{eq:nnUNetdiceloss}
\end{equation}
However, due to the applied patch-based training, we propose to used the class adaptive Dice loss formulation in combination with the basic CE loss. We define the class adaptive Dice loss as:
\begin{equation}
    \text{ca-Dice}(P,G)=\frac{1}{N}\sum_{b,c} \frac{2 \sum_{v}P_{bcv}G_{bcv}}{\sum_{v}P_{bcv}+\sum_{v}G_{bcv}+\epsilon}, \;   
    N = \sum_{b,c} 
    \begin{dcases}
        0,& \text{if} \sum_{v}G_{bcv}=0\\
        1,& \text{else}
    \end{dcases}
    \label{eq:la_diceloss2}
\end{equation}
Differently to the original Dice loss our definition only involves the $N$ classes present in the sampled patch and thus evaluates to the real DSC of the sampled patch instead of considering missing classes as perfectly segmented, which biases the loss towards incorrect scores.

\section{Results}
\label{li:sec:exp_res}

The nnU-Net as a general purpose segmentation framework is based on a fixed and a dataset dependent set of parameters. The patch-size defining rule of the network is based on the assumption that large windows have a more global context and hence improve the segmentation result. However, using the standard deviation $\sigma$ of the organ volume ratios as a cost function to optimize the class imbalance within the patches, results in smaller patch-sizes than the global context maximizing patch-size assumption of the nnU-Net. Our measurement of the class ratio standard deviation $\sigma$ naturally shows that the class imbalance is maximal ($\sigma=0.3301$) if a whole image approach is used and minimal if the patch-size is minimal ($\sigma=0.27146$ for a patch-size of $8\times8\times8$). Figure~\ref{li:fig:li} shows the organ volume ratios, including the background of the sampling process using four different patch-sizes. 

In our experiments, to investigate the effect of the patch-size and thus the class imbalance on the segmentation quality we use the suggested patch-size of the nnU-Net framework and half of the patch-size in-plane and a slightly reduced size out-plane to still give the network enough context in axial direction, resulting in the small patch-size $96\times80\times48$ ($\sigma=0.32337$). We omit the full volume strategy presented in Figure~\ref{li:fig:li} as being infeasible due to its GPU memory demands as well as the minimal possible patch-size only allowing a shallow U-Net with one downsampling (upsampling) layer. Additionally, we include our class adaptive Dice loss formulation (Section~\ref{li:sec:la_dice_loss}) into the nnU-Net loss, as a robust cost function for the patch-based training of datasets with sparse class distributions as given in the HAN area. Consequently, we conduct experiments with the original nnU-Net parameters (large patch-size $192\times160\times56$, nnU-Dice+CE loss) and our introduced class imbalance optimized patch-size and the class adaptive loss function (ca-Dice). 

\begin{figure}
    \centering
    \includegraphics[width=0.99\textwidth]{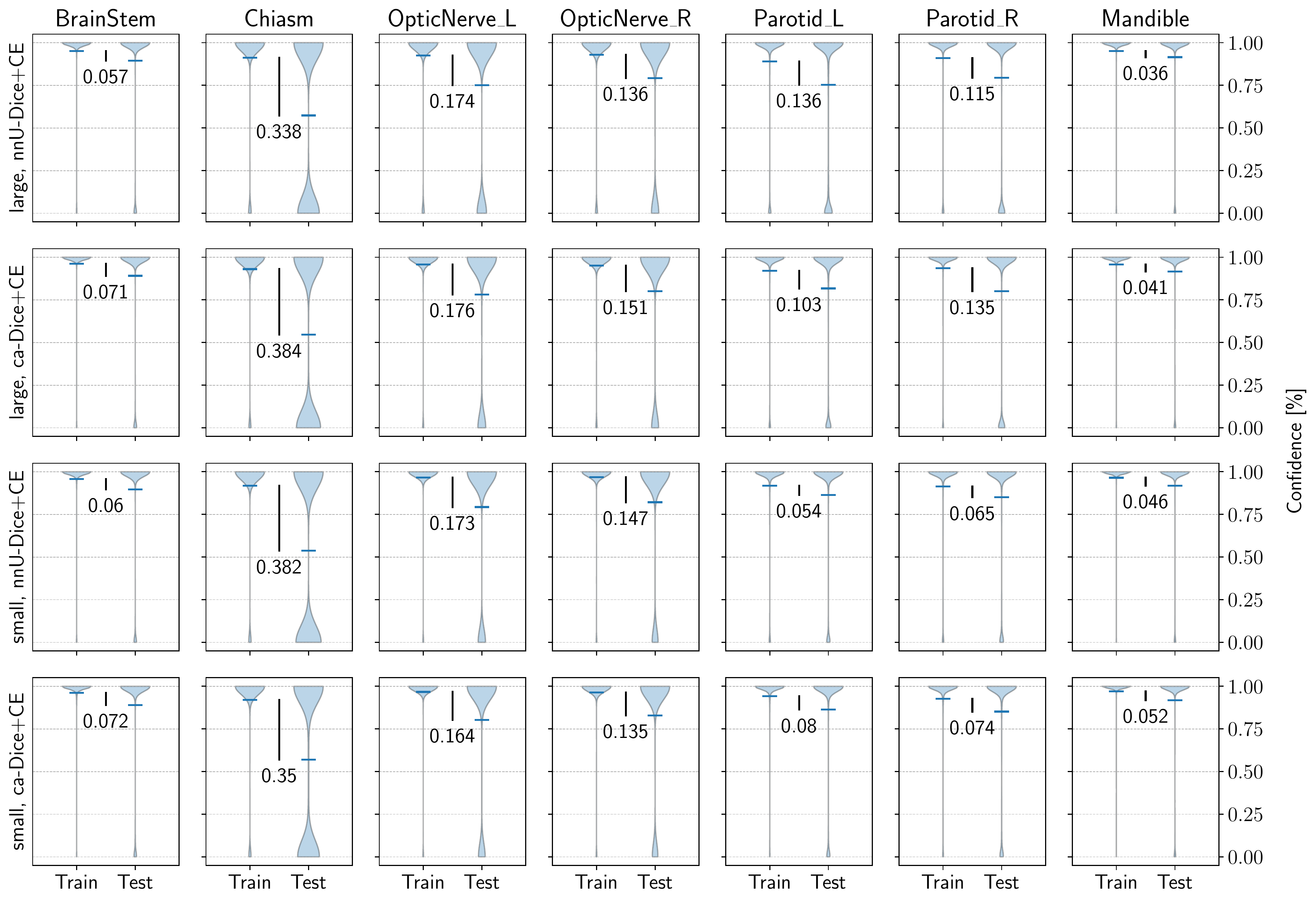}
    \caption{Violin plot of the output confidence distribution of the training and the test samples for the segmented organs of our experiments, with the distance of the average confidence from the training to the test data indicating the potential of overfitting.}
    \label{li:fig:overf}    
\end{figure}

Table~\ref{li:table:avg_res} shows the average results of the networks trained on the MICCAI 2015 HAN challenge dataset~\cite{Raudaschl2017}, according to the challenge protocol. The results on the test data are evaluated using the DSC, the 95HD as well as the surface Dice (SD) as introduced by Nikolov et al.~\cite{Nikolov2018} combining a volume and a surface-based measurement (with surface tolerance $\tau$ identified by the authors in their observer agreement study). The bold values indicate the best results for the given measurement and values marked with stars significance
(Wilcoxon signed rank test with $p < 0.05$) over the baseline. Following the work of Li et al.~\cite{Li2021} in order to analyze a potential overfitting of the small organs we present in Figure~\ref{li:fig:overf} a comparison of the output confidence distribution of the training and the test samples for the segmented organs of our experiments as violin plot. The values in each plot indicate the distance of the average confidence from the training to the test data. 

\begin{table}
    \centering
    \caption{Segmentation results on the combined on- and off-site test data of the MICCAI 2015 HAN challenge dataset~\cite{Raudaschl2017}, for the evaluated configurations in terms of DSC, 95HD and surface Dice (SD). Bold values indicate the best results for the respective organ in each column and values marked with stars significance (Wilcoxon signed rank test with $p<0.05$) over the baseline.}
    \label{li:table:avg_res}
    \begin{tabular}{@{} l l D{?}{\,\pm\,}{3.3} D{?}{\,\pm\,}{4.5} D{?}{\,\pm\,}{3.4}@{}}
        \toprule
        \multicolumn{1}{l}{\textbf{Config (patch-size, loss)}} &
        \multicolumn{1}{l}{\textbf{Organ}} & 
        \multicolumn{1}{c}{\textbf{DSC}} & \multicolumn{1}{c}{\textbf{95HD [mm]}} & \multicolumn{1}{c}{\textbf{SD}} \\
        \midrule
        \midrule
        \multirow{7}{*}{large, nnU-Dice+CE} &    BrainStem & 0.88 ? 0.02 &    3.29 ? 0.67 & 0.96 ? 0.03 \\
         & Optic Chiasm & 0.54 ? 0.21 &    3.48 ? 2.02 & 0.84 ? 0.25 \\
         &     Mandible & 0.94 ? 0.01 &    2.13 ? 1.04 & 0.91 ? 0.05 \\
         & OpticNerve\_L & 0.68 ? 0.11 &    4.91 ? 3.95 &  0.92 ? 0.10 \\
         & OpticNerve\_R &  0.70 ? 0.08 &     3.10 ? 2.46 & 0.96 ? 0.06 \\
         &    Parotid\_L & 0.82 ? 0.08 &    5.36 ? 2.45 & 0.91 ? 0.06 \\
         &    Parotid\_R & 0.84 ? 0.11 &    6.07 ? 4.77 &  0.91 ? 0.10 \\
        \midrule
        \multirow{7}{*}{large, ca-Dice+CE} &    BrainStem & 0.88 ? 0.02 &    3.16 ? 0.45 & 0.96 ? 0.06 \\
          & Optic Chiasm & 0.53 ? 0.21 & 69.42 ? 257.44 & 0.85 ? 0.25 \\
          &     Mandible & 0.94 ? 0.01 &    1.86 ? 0.65 & 0.91 ? 0.05 \\
          & OpticNerve\_L & 0.72 ? 0.08 &    \textbf{2.82} ? \textbf{2.06} & \textbf{0.97} ? \textbf{0.05} \\
          & OpticNerve\_R &  0.70 ? 0.07 &  \textbf{2.21} ? \textbf{0.45} & \textbf{0.99} ? \textbf{0.01} \\
          &    Parotid\_L & 0.86 ? 0.04 & 4.43 ? 1.62 & 0.94 ? 0.04 \\
          &    Parotid\_R & 0.83 ? 0.12 &    5.83 ? 5.27 & 0.91 ? 0.12 \\
         \midrule
         \multirow{7}{*}{small, nnU-Dice+CE} &    BrainStem & 0.88 ? 0.02 &    \textbf{3.13} ? \textbf{0.63} & \textbf{0.96} ? \textbf{0.02} \\
           & Optic Chiasm & 0.53 ? 0.21 &    \textbf{3.23} ? \textbf{1.22} & \textbf{0.89} ? \textbf{0.13} \\
           &     Mandible & 0.94 ? 0.02 &    1.74 ? 0.75 & 0.92 ? 0.04 \\
           & OpticNerve\_L & 0.71 ? 0.07 &    3.03 ? 2.08 & 0.96 ? 0.06 \\
           & OpticNerve\_R & \textbf{0.73} ? \textbf{0.05} &    2.29 ? 0.48 & 0.98 ? 0.02 \\
           &    Parotid\_L & 0.88 ? 0.02 &    4.34 ? 2.44 & 0.95 ? 0.03 \\
           &    Parotid\_R & \textbf{0.88} ? \textbf{0.02} &    4.24 ? 1.61 & 0.93 ? 0.05 \\
          \midrule
          \multirow{7}{*}{small, ca-Dice+CE} &    BrainStem & 0.88 ? 0.02 &  3.33 ? 0.69 & 0.96 ? 0.06 \\
            & Optic Chiasm &  \textbf{0.55} ? \textbf{0.2} &    3.38 ? 1.86 & 0.87 ? 0.19 \\
            & Mandible & 0.94 ? 0.02 &    \textbf{1.72} ? \textbf{0.69} & 0.92 ? 0.04 \\
            & OpticNerve\_L & \textbf{0.73} ? \textbf{0.08} &    2.86 ? 2.13 & 0.97 ? 0.06 \\
            & OpticNerve\_R & 0.72 ? 0.07 &    2.53 ? 1.45 & 0.98 ? 0.03 \\
            & Parotid\_L & \textbf{0.88} ? \textbf{0.02} &  \textbf{4.27} ? \textbf{1.83} & \textbf{0.95} ? \textbf{0.02} \\
            &    Parotid\_R & \textbf{0.87} ? \textbf{0.04} & \textbf{4.08} ? \textbf{1.39} & \textbf{0.94} ? \textbf{0.04} \\
        \midrule
        \midrule
        large, nnU-Dice+CE & average & 0.77 ? 0.17 &  4.05 ? 3.05 & 0.92 ? 0.12 \\
        large, ca-Dice+CE & average & 0.78 ? 0.16 & 12.82 ? 97.3\textsuperscript{$\star$} & 0.93 ? 0.11\textsuperscript{$\star$} \\
         small, nnU-Dice+CE & average & 0.79 ? 0.16 &  \textbf{3.14} ? \textbf{1.69}\textsuperscript{$\star$} & \textbf{0.94} ? \textbf{0.07}\textsuperscript{$\star$} \\
         small, ca-Dice+CE   & average &  \textbf{0.80} ? \textbf{0.15}\textsuperscript{$\star$} &   3.17 ? 1.70\textsuperscript{$\star$} & 0.94 ? 0.09\textsuperscript{$\star$} \\
        \bottomrule 
    \end{tabular}
\end{table}

\subsection{Implementation Details}
Our implementation is based on the Monai DynUNet pipeline module\footnote{\url{https://github.com/Project-MONAI/tutorials/} (accessed 2021-12-21)}, a reimplementation of the dynamic UNet used in the nnU-Net framework~\cite{Isensee2021} and further adapted to follow the nnU-Net parameterization. Monai is a PyTorch-based framework for deep learning in healthcare imaging\footnote{\url{https://monai.io/} (accessed 2021-12-21)}. Our models are trained on Nvidia Titan RTX GPUs with 24 GB of memory for an average of 67 hours.

\section{Discussion}
\label{li:sec:dis_con}
The results of our experiments in Table~\ref{li:table:avg_res} reveal that the presented extensions to the nnU-Net framework, the patch-size adjustment especially in conjunction with the class adaptive Dice loss, are favorable for the present class imbalance in the HAN area. 

Reducing the patch-size directly influences the class imbalance within the sampled patches. The standard deviation,  introduced as a measurement for the volume ratio imbalance within a training image patch, changes from $\sigma=0.32605$ to $\sigma=0.32337$ using the GPU memory optimized large patch-size of $192\times160\times56$ compared to the suggested class imbalance optimized small patch-size of $96\times80\times48$. As visible in Figure~\ref{li:fig:li}, especially the ratio of the smaller classes increases within a patch. The improvement of the class imbalance therefore reduces the bias towards the large classes during the training and effectively results in an increase in performance of 2\% in terms of the DSC and a significant increase of 2\% regarding SD compared to the baseline nnU-Net framework. The 95HD is significantly reduced by 0.91 mm, yielding an improvement of 22\% compared the baseline.

The utilization of the class adaptive Dice loss in the loss formulation of the nnU-Net improves the segmentation results regarding the DSC and significantly the SD by another 1\%. The average of the 95HD is not improved as the Optic Chiasm is not segmented in one test sample, however all other single organ measurements show improvements over the baseline. Contrary to the standard multi-class Dice loss formulation the class adaptive Dice loss only evaluates the classes available within each patch, whereas the standard Dice loss calculates the average over all classes, distorting the average DSC depending on the current network prediction of the missing classes. The nnU-Dice which is based on the batch-Dice formulation~\cite{Kodym2018}, however, reduces the risk of missing classes by considering the batch dimension as part of the patch volume. The risk of missing classes within a patch depends on the volume size, the class distribution within the whole volume and, as adjustable training parameters, the patch-size and the sampling strategy. As the nnU-Net framework uses a 33\% random foreground oversampling strategy, the large patches and the batch-Dice formulation make the baseline nnU-Net already stable against missing classes. Nonetheless, we argue to use the class adaptive Dice, as it is robust against missing classes, especially if the patch-size is smaller and the class distribution within the volume sparse. By showing significantly improved segmentation results for all measures our experiments support the usage of a combined small patch-size and the class adaptive Dice for imbalanced segmentation problems. 

Deviating from the suggestion of the original work of Li et al.~\cite{Li2021}, Figure~\ref{li:fig:overf} does not show the direct network output (the logits) of the segmented classes and its corresponding decision boundaries, which is only possible for up to three classes, but the confidence distribution after the softmax normalization of the eight HAN organs to be segmented. Although no decision boundary can be depicted for more then three classes, the presentation of the organ-wise normalized confidence values allows a direct comparison of the average confidence drift from train to test time and thus the identification of overfitting. The results in Figure~\ref{li:fig:overf} confirm the findings of Li et al.~\cite{Li2021} for the class imbalanced HAN area and show that the small organs (the Optic Chiasm and the Optic Nerves) are subject to larger differences in training and test time confidence and accordingly prone to overfitting. The measurements also indicate the overall performance enhancement of the ca-Dice loss over the baseline, visible in the increased average confidence values, but do not show a reduction of the overfitting of the small organs by the loss function adaption. Contrary, the experiments with the small patch-size optimized to reduce the class imbalance, show a clear average confidence difference reduction of the Parotid Glands. The Parotid Glands can be considered as mid-sized organs, allowing the assumption that a further reduction of the class imbalance can reduce the confidence drift for the small organs too and hence increase their final segmentation results. The assumption is also supported by the constantly small average confidence drift of the Mandible and the Brainstem being the largest organs with the largest patch ratio and consequently the least overfitting. 

Finally, in Table~\ref{li:table:compare} we present the segmentation results combining the small patch-size and the class adaptive Dice and chronologically compare them with the segmentation results of the most important works in the field also presenting their results on the MICCAI 2015 HAN challenge dataset. The Table also indicates the number of organs and data samples used, as the original challenge protocol and its defined data splits are not followed in general.

\begin{table}[]
    \centering
    \caption{Average DSC and 95HD on the MICCAI HAN challenge dataset}
    \label{li:table:compare}
    \begin{tabular}{@{} l D{?}{\,\pm\,}{3.3} D{?}{\,\pm\,}{3.3} c c}
        \toprule
        \multicolumn{1}{l}{\textbf{Literature}} &
        \multicolumn{1}{c}{\textbf{DSC}} & \multicolumn{1}{c}{\textbf{95HD [mm]}} & \multicolumn{1}{c}{\textbf{Data}} & \multicolumn{1}{c}{\textbf{Organs}} \\
        \midrule
        Raudaschl et al.\cite{Raudaschl2017} & 0.76 &  - & 25/15 & 9 (2 partly) \\
        Fritscher et al. \cite{Fritscher2016} & 0.66 ? 0.08 &  - & 20/10 & 6 \\
        Tappeiner et al.\cite{Tappeiner2019} & 0.72 ? 0.18 &  6.30 ? 16.2 & 25/15 & 7 \\
        Zhu et al. \cite{Zhu2019} & 0.79 ? 0.05 &  - & 38/10 & 9 (3 partly) \\
        Tappeiner et al. \cite{Tappeiner2020} & 0.75 ? 0.16 & 3.02 ? 1.92 & 25/15 & 7  \\
        Guo et al. \cite{Guo2020} & 0.82 ? 0.05 &  - & 33/15 & 9 (3 partly) \\
        Gao et al. \cite{Gao2020} & 0.85 ? 0.06 & 2.17 ? 0.93 & 38/10 & 9 (3 partly) \\
        Nikolov et al. \cite{Nikolov2018}   & 0.81 ? 0.05 &  - & (663)/15 & 8 (2 partly) \\
        Chen et al. \cite{Chen2021} & 0.81 ? 0.05 &  - & 33/10 & 9 (3 partly) \\
        Tang et al. \cite{Tang2021} & 0.83 ? 0.05 &  - & 33/15 & 9 (3 partly) \\
        our (small, ca-Dice+CE) & 0.80 ? 0.15 &  3.17 ? 1.69 & 25/15 & 7 \\ 
        
        \bottomrule 
    \end{tabular}
\end{table}

\section{Conclusion}
In summary, in this work we present an intuitive measurement for the organ volume ratio difference, which is a central problem appearing in the DL based segmentation of the HAN area. Based on the measurement we optimize the patch-size parameter regarding the class imbalance for a single network based HAN segmentation architecture. Additionally, we utilize the class adaptive Dice as a robust loss function for missing classes within a training patch. Both adaptions are incorporated in the nnU-Net framework where we are able to increase the segmentation results by an additional 3\% in terms of the DSC and the SD and by 22\% regarding the 95HD, resulting in an average DSC of $0.8\pm0.15$ and a 95HD of $3.17\pm1.7$ mm for the segmented HAN organs respectively. 

The patch-size optimization and the class adaptive Dice loss can both easily be integrated into current DL based segmentation approaches. In future work we want to improve the state-of-the-art performance of the recently presented hybrid 2D-3D, single network approach of Chen et al.~\cite{Chen2021} by integrating our adaptations. Single network approaches are end-to-end trainable, less complex and therefore of higher practical interest compared to complex multi-network solutions. As an addition to the overfitting analysis we like to investigate and combine asymmetric loss functions therms, proposed by Li et al.~\cite{Li2021} with our ca-Dice loss to increase the distance to the decision boundaries of the small classes to further increase their test time performance. 

\section*{Declarations}

\subsection*{Conflict of interest}
Elias Tappeiner, Martin Welk, Rainer Schubert declare to have no conflict of interest.

\subsection*{Ethical approval}
All procedures performed in studies involving human participants were in accordance with the ethical standards of the institutional and/or national research committee and with the 1964 Helsinki declaration and its later amendments or comparable ethical standards.

\subsection*{Informed Consent}
Patients informed consent was given in the data originating clincal study~\cite{Ang2014}.

\subsection*{Availability}
The anonymized MICCAI 2015 HAN challenge dataset is publicly available (\url{http://www.imagenglab.com/newsite/pddca/}). The code of the work is available on Github (\url{https://github.com/elitap/classimbalance}).

\bibliographystyle{unsrt}  
\bibliography{references}

\end{document}